\documentclass[twoside, twocolumn]{article}

\usepackage[english]{babel}

\usepackage[sc]{mathpazo}

\usepackage[T1]{fontenc}
\usepackage{microtype}
\usepackage[margin=2cm, hmarginratio=1:1, top=32mm, columnsep=20pt]{geometry}
\usepackage[hang, small, labelfont=bf, up, textfont=it, up]{caption}
\usepackage{booktabs}
\usepackage{lettrine}

\usepackage{amsmath}

\usepackage{enumitem}
\setlist[itemize]{noitemsep}

\usepackage{abstract}

\usepackage{titlesec}
\renewcommand\thesection{\Roman{section}}
\renewcommand\thesubsection{\roman{subsection}}
\titleformat{\section}[block]{\large\scshape\centering}{\thesection.}{1em}{}
\titleformat{\subsection}[block]{\large}{\thesubsection.}{1em}{}

\usepackage{fancyhdr}
\usepackage{titling}
\usepackage{hyperref}
\usepackage{subcaption}
\usepackage{dblfloatfix}
\usepackage{floatrow}
\usepackage{multirow}
\usepackage{graphicx}
\graphicspath{ {images/} }

\pretitle{\begin{center}\Huge\bfseries}
\posttitle{\end{center}}
\title{Fast and Accurate Person Re-Identification with RMNet}
\author{%
\textsc{Evgeny Izutov} \\[1ex]
\normalsize IOTG Computer Vision (ICV), Intel \\
\normalsize \href{mailto:evgeny.izutov@intel.com}{evgeny.izutov@intel.com}
}
\date{} 
\renewcommand{\maketitlehookd}{
\begin{abstract}
\noindent In this paper we introduce a new neural network architecture designed to use in embedded vision applications. It merges the best working practices of network architectures like MobileNets and ResNets to our named RMNet architecture. We also focus on key moments of building mobile architectures to carry out in the limited computation budget. Additionally, to demonstrate the effectiveness of our architecture we evaluate the RMNet backbone on  Person Re-identification task. The proposed approach is in top 3 of state of the art solutions on Market-1501 challenge, however our method significantly outperforms them by the inference speed.
\end{abstract}
}

\begin{document}

\maketitle


\section{Introduction}

\lettrine[nindent=0em,lines=3] {T}{}he CNN-based solutions have demonstrated the ability to solve a wide range of computer vision tasks achieving the human level performance or even outperforming them. Moreover, not only demonstrating the ability to solve a set of canonical tasks like ImageNet \cite{ILSVRC15} classification or Cityscapes \cite{Cordts2016Cityscapes} segmentation challenges can be attributed to CNNs but solving practical use case problems. It's about issues like a person re-identification which  is a key component of the tracking pipelines.

\par Unfortunately, many researchers offering each time a dramatically new approach allowing to lift a problem on the new level of understanding have a purpose of their work to only beat  current state of the art without any attention to the performance problem. But speaking about the industry-useful solutions we should take into account the requirement of real-time inference on the customer affordable hardware.

\par In case of CNN-based solutions the necessity to affect the inference behavior the choice of backbone is the only thing that needs to be changed. We have many examples of backbone architectures like MobileNet (\cite{DBLP:journals/corr/HowardZCKWWAA17}, \cite{DBLP:journals/corr/abs-1801-04381}) and ShuffleNet (\cite{DBLP:journals/corr/ZhangZLS17}, \cite{Ma_2018_ECCV}) designed for the fast inference in embedded applications. The most significant moment is that for many users these backbones are the only changes required to adopt their approach for the fast inference. Instead of thinking in terms of practices satisfying their target requirements, users mix the components from different and often incompatible areas and, as result, underperform what it could be.

\par In this paper, we address this issue by carefully designing the direct architecture to solve specific and small task like a person re-identification. Our aim is to show that this problem can be solved on near state of the art level and significantly outperformed by speed. Our contributions are as follows:
\begin{itemize}
\item New lightweight backbone architecture named RMNet for the fast and accurate inference for mobile applications.
\item Re-thinking of the manifold learning techniques according to the person re-identification challenge.
\item Novel lightweight network head to combine the advantages of the low and high level losses without grow in number of parameters.
\end{itemize}

\par More broadly, this work demonstrates some ways to design the lightweight CNN-based solution to tackle with specific (not general) tasks without needs to accept being fast as well as being is inaccurate. The proposed model (set of models with different trade off between speed and accuracy) you can find as a part of the Intel$^{\textregistered}$ OpenVINO\texttrademark toolkit\footnote{For more information you can follow the link: \href{https://software.intel.com/en-us/openvino-toolkit}{https://software.intel.com/en-us/openvino-toolkit}}.

\par \textbf{Index terms.} Person re-identification, manifold learning, local and global structure losses, mobile network architecture, lightweight backbone, RMNet.


\section{Related Work}

\subsection{Mobile architectures}

\lettrine[nindent=0em,lines=3]{I}{}n recent time Deep Learning (DL) as an independent tool of Machine Learning has made significant leap in a CNN architecture development starting from the vanilla networks like VGG \cite{DBLP:journals/corr/SimonyanZ14a} and continue with ResNet \cite{DBLP:journals/corr/HeZRS15} and Inception \cite{DBLP:journals/corr/SzegedyLJSRAEVR14} families. Recent architectures bring some key understanding at how to deal with permanent DL problems: vanishing/exploding gradients, over-parametrization and next over-fitting. As a byproduct for the fast inference purposes we get some reduction in the computation budget while using ResNet-18 and similar models which we can name "relatively small". For some simple tasks the cheap speed up of inference by reduction of the depth of some default architecture is enough and no future investigation is performed on this aspect. But when we speak about mobile applications the future inference time reduction is needed. On the way to do that the techniques like model weights pruning \cite{DBLP:journals/corr/abs-1801-07365} and quantization \cite{2018arXiv180608342K} are used.

\par The first one is based on the assumption that the trained CNN-based model has a parameter redundancy \cite{DBLP:journals/corr/ChengYFKCC15} by some imperfection of the Stochastic Gradient Descent (SGD) based training procedure which sins to produce duplicate filters \cite{RoyChowdhury2018ReducingDF}. The main idea of pruning methods is to remove useless parameters without significant drop in accuracy. As it can be seen, the recent papers demonstrate model compression and inference speed up pretty well \cite{DBLP:journals/corr/abs-1801-07365}. But the parameter redundancy problem has another point of view -- instead of putting up with the necessity to use pruning we can try to train a model directly without any parameter redundancy. In the proposed paper we have investigated one of possible ways to get it.

\par Regarding a quantization or more restricted binarization \cite{DBLP:journals/corr/CourbariauxB16} techniques we do not consider this issue because it's mostly related to edge-specific implementations than general ideas which are applicable for the wide range of tasks.

\par Completely different approach is to design the network architecture directly assuming some possible degradation in the accuracy but with gain in a computation time. The first significant step by introducing the depth-wise separable convolutions \cite{DBLP:journals/corr/HowardZCKWWAA17} has been made. This idea was simple but powerful. In present time all mobile network architectures reuse it including the proposed paper too. To future speed up the computations the MobileNet-v2 \cite{DBLP:journals/corr/abs-1801-04381} architecture focuses on an idea of fixing some internal problems with ReLU \cite{Nair:2010:RLU:3104322.3104425} activation function by inverting well-known bottlenecks. We agree with authors that some problems arise because of incompatibility SGD properties with  ReLU function. But we have found out that refusal in favor of ELU \cite{DBLP:journals/corr/ClevertUH15} activation function with some other changes in backbone is more flexible way.

\par A special place occupies the ShuffleNet \cite{DBLP:journals/corr/ZhangZLS17} architecture which brings an idea to utilize multipath inference in single network by channel shuffling. Next generation \cite{Ma_2018_ECCV} of this architecture optimizes the the memory consumption and reveals some analogies with DenseNet \cite{DBLP:journals/corr/HuangLW16a} architecture.

\subsection{Person re-identification}

\lettrine[nindent=0em,lines=3]{T}{}he person re-identification task is formulated as a task of learning of some parametric mapping function $f_{\theta}: R^F \mapsto R^D$ which maps semantically similar points from the image space $R^F$ onto close points on the embedding space $R^D$. During the inference a pair of input images is compared by $L_2$ or cosine distance between the embeddings vectors.

\par For now the best working practices utilize the Siamese network \cite{Chopra:2005:LSM:1068507.1068961} with appropriate target function like the triplet loss \cite{DBLP:journals/corr/SchroffKP15} as well as train a model as a classification task with Softmax and cross-entropy loss \cite{DBLP:journals/corr/JanochaC17}. More recently they reuse the AM-Softmax loss \cite{DBLP:journals/corr/abs-1801-05599} from the twin face recognition challenge.

\par Next improvement in person re-identification has been connected with joint training both metric learning approaches (triplet and AM-Softmax losses), incorporating some form of attention by slicing images on horizontal stripes \cite{DBLP:journals/corr/abs-1711-09349}, aggregation of embeddings from different levels \cite{DBLP:journals/corr/abs-1804-05275} and mix of the previous attempts in single network without regard for the computation budget \cite{DBLP:journals/corr/abs-1804-01438}.

\par Another attempt to resolve the person re-identification challenge is based on some kind of hard sample mining techniques for both the triplet loss and for joint training \cite{DBLP:journals/corr/HermansBL17}.

\par Regarding the presented paper we are focused on manual mixing different metric learning approaches to escape the difficulties of triplet sampling and incorporate different-level manifold learning (\cite{10.1007/978-3-319-46478-7_31}, \cite{DBLP:journals/corr/abs-1804-03864}).

\section{Backbone design}

\begin{figure*}[th]
\centering
\includegraphics[width=0.9\textwidth]{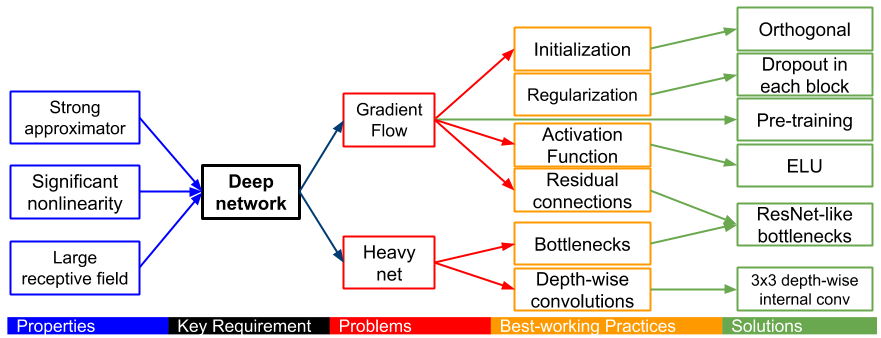}
\caption{Thoughts-flow diagram to build the target lightweight architecture by the definition of key requirements and solving the following issues.}
\label{fig:top-down-view}
\end{figure*}

\subsection{Top-Down architecture design}
\lettrine[nindent=0em,lines=3]{A}{s} it was previously sad, the evolution of network architectures has made several steps on the way from the regular structure where the representation power is focused in simple stacking of convolution layers to architectures which exploit the fusion of different-level representations into a single stage. The last trend is to concentrate on the network design in variation of its building blocks like bottlenecks in ResNet architecture. This strategy is followed by the recent mobile architectures: MobileNet and ShuffleNet.

\par Regarding the design of a network for mobile applications, we can follow one of possible approaches. The first one is a "bottom-up" approach which is based on discovering the inference bottlenecks and following fixes for them. The most powerful example of such approach is ShuffleNet-v2 \cite{Ma_2018_ECCV} architecture. It includes strong baseline to exclude as much memory consumed operations as possible. Generally speaking it's a good attempt to build a fast network foremost but without any attention to the target task. Final accuracy in this case is mostly a result of lucky choice of architecture, otherwise the incrementation of the model size is proposed only.

\par Another approach is presented by a "top-down" one. It includes the definition of key requirements which cannot be omitted and the following growing of the network building blocks. Moreover, such requirements don't need to be of one and the same logical level. Often this list is composed of high-level architecture solutions (shallow or deep network) and low-level operations. All the next steps are targeted to merge requirements into a single multi-level solution. It is worth saying that next steps are not limited in an architecture design only but may include initialization tricks and more sophisticated training procedure.

\par Of course it may happen that key requirements supplemented by the limited computation budget are contradictory. Fortunately we should remember that our purpose is not focused on developing the solution for some general task (e.g. ImageNet \cite{ILSVRC15} classification or COCO \cite{DBLP:journals/corr/LinMBHPRDZ14} detection problems). It gives the hope that the relief in generality of model brings us the realizable trade-off. In this paper we follow the "top-down" approach and  the next sections show our vision on direct building of a lightweight model according to the specific person re-identification task.

\subsection{Deep vs Shallow networks}

\lettrine[nindent=0em,lines=3]{I}{}n the course of the conversation about a network design in the limited computation budget we face well known dilemma of deep or shallow network architecture. Most often the choice is to cut a general architecture to satisfy the restriction on maximal number of FLOPs per the single network input. In the architecture level it means the aggressive usage of pooling operations on the early stages (e.g. \cite{Ma_2018_ECCV}). On the one hand, the pooling operator should bring some kind of transformation which is the equivariant to translations. Unluckily, for the rest of the tasks aggressive pooling prevents from extracting of accurate high-level features.

\par On the other hand, we cannot give up pooling operators because it is a lightweight way to control the number of FLOPs on each scale level by changing the spatial resolution of the feature map. In addition to that, we can vary the number of blocks on each scale and the width of each block. Unfortunately, for most of users the restriction of number of blocks without any change in each of them is the easiest way.

\begin{figure}[b]
\includegraphics[width=0.9\textwidth]{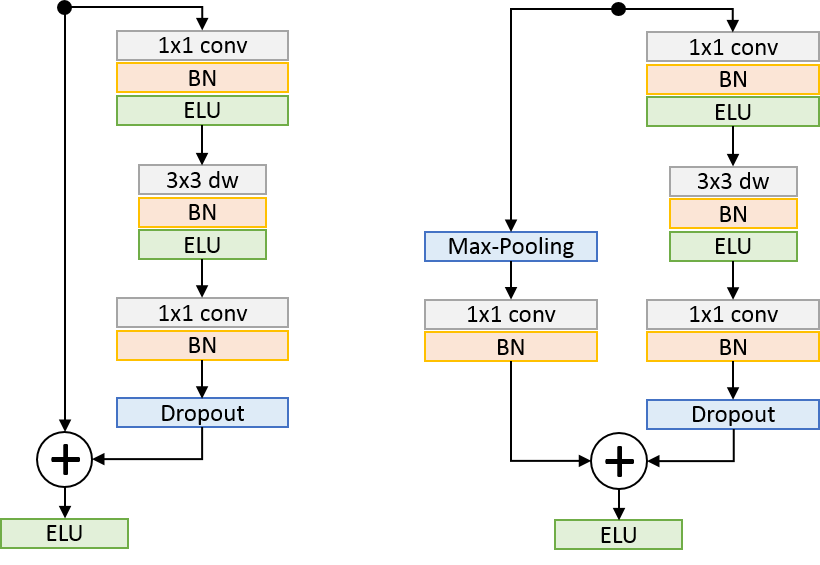}
\caption{Diagram of RMNet block. Left: regular bottleneck. Right: bottleneck for spatial reduction with stride 2 for max-pooling and internal convolution layers.}
\label{fig:rm-blocks}
\end{figure}

\par In the presented paper we defend the position that the key component for robust feature extractor is the depth of a network (in terms of number of convolutions in the longest path from the input to the network output). Regarding the design of a backbone the choice is imposed on a fight with a gradient flow during training and is based on the ResNet-family. According to the results \cite{DBLP:journals/corr/VeitWB16} the residual structure of bottlenecks can be interpreted as an iterative feature enhancing on a single representation level (obviously the level border is defined by down-scaling operations). It is also important to note that ResNet-18 or even ResNet-50 architectures are too "shallow" and don't satisfy our intuition. We should talk about a hundred of layers at least.

\par But choosing the deep architectures we face the necessity to make each residual block as light as possible. The simplest way is to follow the mainstream practice to use bottlenecks with two consecutive $3 \times 3$ convolutions instead of original $1 \times 1 \rightarrow 3 \times 3 \rightarrow 1 \times 1$ \cite{DBLP:journals/corr/HeZR016}. Contrariwise, we can think about a network depth in terms of the representation power \cite{DBLP:journals/corr/WangRX17}. For us it means that the choice to use either two or three convolutions in the bottleneck is decided in favor of three convolutions with non-linearity after each.

\par Finally, we can formulate the list of key requirements which forms the basis of the presented backbone architecture (on Figure \ref{fig:top-down-view} you can see our flow of thoughts on the way to build lightweight network):
\begin{itemize}
\item Very deep network with a hundred of layers.
\item ResNet-like architecture.
\item Residual blocks with three convolutions ($1 \times 1 \rightarrow 3 \times 3 \rightarrow 1 \times 1$) and non-linearity after each.
\end{itemize}

\subsection{RMNet backbone}

\lettrine[nindent=0em,lines=3]{F}{}or now we have the general vision on a backbone design and support points to fit a model to the target computation budget. As it was mentioned earlier the ResNet-like bottlenecks consist of 3 convolutions: the first $1 \times 1$ convolution maps the input onto some internal representation with simultaneous reduction of number of channels, the next internal $3 \times 3$ convolution carry out spatial mixing and the last $1 \times 1$ convolution maps internal representation back onto the input manifold.

\begin{table}[h]
\begin{tabular}{l|c|c|c|c}
\multicolumn{1}{c|}{\textbf{Name}} & \textbf{Times} & \multicolumn{1}{l|}{\textbf{Stride}} & \textbf{\begin{tabular}[c]{@{}c@{}}Spatial\\ size\end{tabular}} & \textbf{\begin{tabular}[c]{@{}c@{}}Num\\ channels\end{tabular}} \\ \hline
Input &  &  & $\times 1$ & 3 \\ \hline
$3 \times 3$ conv & 1 & 2 & $\times 1/2$ & 32 \\ \hline
RM-block & 4 & 1 & $\times 1/2$ & 32 \\
RM-block & 1 & 2 & $\times 1/4$ & 64 \\
RM-block & 8 & 1 & $\times 1/4$ & 64 \\
RM-block & 1 & 2 & $\times 1/8$ & 128 \\
RM-block & 10 & 1 & $\times 1/8$ & 128 \\
RM-block & 1 & 2 & $\times 1/16$ & 256 \\
RM-block & 11 & 1 & $\times 1/16$ & 256
\end{tabular}
\caption{RMNet backbone architecture}
\label{table:backbone}
\end{table}

\par The first step to reduce the number of operations is to replace the internal $3 \times 3$ convolution with its depth-wise variant \cite{DBLP:journals/corr/HowardZCKWWAA17}. But instead of the depth-wise separable convolution practice \cite{DBLP:journals/corr/Chollet16a} we preserve the nonlinearity after the internal convolution to leave unchanged the representation power of the network. Unfortunately, this reduction is not enough and the last support point should be used too. This is about the channel reduction factor used in the internal convolution. In this paper we need to use strong $\times 1/4$ factor. Moreover the maximal number of channels is also limited 256 too.

\par Another unobvious question is about the choice of an activation function. The common practice is to use ReLU \cite{Nair:2010:RLU:3104322.3104425} non-linearity. It is found out that  some negative effect of using ReLU in deep networks (\cite{DBLP:journals/corr/abs-1801-04381}, \cite{DBLP:journals/corr/HeZR016}) which is connected with well known sparsity of activations. Easy to see that this sparsity in forward pass will affect the backward pass too by producing sparsity in gradients and the following convergence retardation. Researchers propose different solutions but we follow more simple way to replace ReLU onto ELU \cite{DBLP:journals/corr/ClevertUH15} activation function. As it will be described further it dramatically changes the behavior of the network.

\par The next important question is related to the utilization of model parameters. Looking at the effectiveness of pruning methods \cite{2018arXiv180807471H} we should take into account the fact that not all learnt model parameters are useful according to the target task. In case of general architectures with millions of parameters it is expected behavior but regarding our network design with strong channel reduction it's impossible to leave some rudimentary parameters.

\par To tackle with the above reported issue we follow the common practices like orthogonal weight initialization \cite{DBLP:journals/corr/SaxeMG13} (not for all filters), pre-training with huge general-purpose datasets \cite{DBLP:journals/corr/HuhAE16} and dropout regularization in each bottleneck \cite{DBLP:journals/corr/PaszkeCKC16}.

\par The final RMNet (Residual Mobile Network) block is presented on Figure \ref{fig:rm-blocks} and whole backbone design is reported in Table \ref{table:backbone}.

\section{ReID network}

\begin{figure*}[th]
\centering
\includegraphics[width=0.9\textwidth]{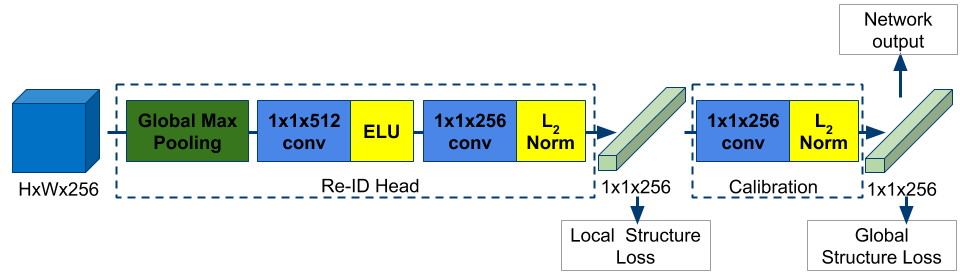}
\caption{Re-identification head to map the internal representation after backbone onto the final embedding vector.}
\label{fig:reid-head}
\end{figure*}

\subsection{Manifold learning}
\lettrine[nindent=0em,lines=3]{A}{}s it was described earlier the goal of the person re-identification based on metric learning is to learn the parametric function $f_{\theta}$ embedding vectors of which can be compared with simple $L_2$ norm. For us it means that learning process can be interpreted as a process of forming the target manifold with desired properties.

\par Generally speaking each loss function impacts different aspects of the final manifold. In light of this we can divide them in two big families: global and local structure losses. Let's describe a set of appearances of different instances. Our goal is to find the transformation after which the appearances of the same instance will be closer to each other rather than to different instances. On the one hand, for this purpose we can select the single appearance (center) of each instance and try to learn mapping by forcing other appearances to be close to its center instance. In other words we define the global rule for the mapping function. And this is the nature of the first family losses. Regarding the examples of implementation there are different modifications of Softmax with Cross-Entropy losses (see eq. \ref{eq:crossentropy}). In the presented paper we are focused on a variant with large margins between classes -- AM-Softmax loss (see eq. \ref{eq:amsoftmax}). \cite{DBLP:journals/corr/abs-1801-05599}.

\begin{equation}
\label{eq:crossentropy}
L_{CE} = - \sum_{i} \log(p_{y_i}), \sum_{j} p_j = 1
\end{equation}

\begin{equation}
L_{glob} = - \sum_{i} \log(\frac{e^{s(W_{y_i}^{T}f_{i}-m)}}{e^{s(W_{y_i}^{T}f_{i}-m)} + \sum_{j, j \neq y_i} e^{sW_{y_j}^{T}f_{i}}})
\label{eq:amsoftmax}
\end{equation}

\par On the other hand, we can follow the Hebbian Learning Rule \cite{10.1007/978-3-642-70911-1_15} which declares that local rules of interactions between elements define the global order of the system. This learning strategy is implicitly presented by the triplet loss \cite{DBLP:journals/corr/SchroffKP15} family. Unluckily, the main drawback of triplets is a sampling procedure which significantly impacts on the final model accuracy \cite{DBLP:journals/corr/HermansBL17}.

\par Recent papers proposed to merge both loss families into a single training procedure and achieved the state of the art results \cite{DBLP:journals/corr/abs-1804-01438}. In our opinion the better performance can be achieved by an elimination of the triplets by dividing them in two constituent forces: push and pull losses \cite{NIPS2016_6200}. In the presented paper we follow the same strategy to divide triples into components thereby overcoming the sampling issues but we supplement the default margins by the "smart" variant like in \cite{DBLP:journals/corr/abs-1804-03864}. Finally we have three local structure losses: Center (eq. \ref{eq:center}), PushPlus (eq. \ref{eq:push}) and GlobPushPlus (eq. \ref{eq:globpush}) losses.

\begin{equation}
\label{eq:center}
L_{center} = \sum_{i} d(f_{\theta}(x_i);c_{x_i})
\end{equation}

\begin{equation}
\label{eq:push}
L_{push} = \sum_{i \neq j} [m + d(f_{\theta}(x_i); c_{x_i}) - d(f_{\theta}(x_i);f_{\theta}(x_j))]_{+}
\end{equation}

\begin{equation}
\label{eq:globpush}
L_{gpush} = \sum_{i} \sum_{\substack{k = \overline{1, C} \\ i \neq k}} [m + d(f_{\theta}(x_i); c_{x_i}) - d(f_{\theta}(x_i);c_k)]_{+}
\end{equation}

\par Total loss to train the model is a weighted sum of global and local losses (weights are estimated to equalize the impact of each loss in the total sum):

\begin{equation}
\label{eq:totalloss}
L = w_{1}L_{glob} + w_{2}L_{center} + w_{3}L_{gpush} + w_{4}L_{push}
\end{equation}

\subsection{Re-identification head}

\lettrine[nindent=0em,lines=3]{T}{}he last component of our network is a re-identification head which maps the point from the internal representation (backbone output) onto the final embedding which can be compared with others by the cosine (or $L_2$) distance. Recently, the unique choice is to use a fully connected (FC) layer on the top of backbone output. Unfortunately, FC layer are too wasteful to the computation resources and cannot be used for mobile networks.

\par Another variant is presented by using global pooling operators like max- or average-pooling. As it is reported in the paper \cite{DBLP:journals/corr/abs-1801-05339} such approach includes some form of the spatial attention due to pooling over all spatial locations of a feature map. We follow the same solution and use global max-pooling (GMP) operator to collapse the spatial dimensions. You can find the proposed re-identification head on Figure \ref{fig:reid-head}.

\par Our re-identification head has two key components. The first one is inverted bottleneck after the GMP operator -- by $1 \times 1$ convolution we increase the number of channels from 256 to 512 and then compress it back to the 256 (attempt to leap in high dimensional space where the class separation can be solved by linear transformation). The second one is based on dividing the support point of global and local structure losses. It means that we extract some internal representation which is trained with local structure losses only and then we calibrate it by learning with global structure losses. For both representations we use $L_2$ normalization to follow the AM-Softmax proposed restrictions on the embeddings (to be compatible with a cosine similarity measure). Finally, the network output is the last calibrated embedding.


\section{Implementation details}

\subsection{Network architecture}
\lettrine[nindent=0em,lines=3]{T}{}he proposed network consists of two consecutive components: lightweight feature extractor (RMNet-based backbone) and single re-identification head. To reduce the total inference time we follow the fully convolutional network (FCN) practice and don't use any FC layers. Moreover we avoid the usage of multibranch \cite{DBLP:journals/corr/abs-1804-01438} solutions and concatenation of embeddings from different layers \cite{DBLP:journals/corr/abs-1804-05275}.

\par The network extracts the $L_2$ normalized embedding vector with 256 elements which can be compared with another one in pairwise manner using the cosine similarity measure.

\subsection{Optimization}
\lettrine[nindent=0em,lines=3]{A}{}ll experiments have been completed in Caffe framework \cite{jia2014caffe}. We use the SGD with momentum optimization method and decay on $10^{-1}$ the learning rate each 50k iteration starting with $10^{-2}$.

\par To initialize the network parameters we use the mixed strategy: input $1 \times 1$ convolutions of each bottleneck are initialized orthogonally \cite{DBLP:journals/corr/SaxeMG13} and the rest weights initialized using MSRA method \cite{DBLP:journals/corr/HeZR015}. Before running the main experiment we pre-trained the backbone on the OpenImages dataset \cite{openimages} by fitting a classification task on the extracted object crops ($224 \times 224$ input size).

\par One more important step to train the lightweight network which is able to utilize significant part of parameters and prevent from the need to use pruning is using dropout regularization \cite{DBLP:journals/corr/abs-1207-0580} in each block (dropout ratio is set to $10\%$). But the dropout regularization reduces the total network capacity and it's unsuitable for our initially small implementation. To overcome this issue we disable the dropout regularization on the late iterations (when the learning rate is small enough) and continue without it. This strategy allows us to form the manifold structure on early iterations without the threat of over-fitting but to use up the whole network capacity later.

\par To solve the unbalanced data problem (significant difference in a number of appearances of each identity) we follow the common practice to reuse the hard sample mining procedure \cite{DBLP:journals/corr/GordoARL16a}. Our implementation of it consists of next steps:
\begin{enumerate}[label=\alph*]
\item To sample $k$ augmented images for each identity from the training dataset.
\item To estimate the value of the loss $w_{1}L_{glob} + w_{2}L_{center} + w_{3}L_{gpush}$ for each sample.
\item To select top $50\%$ of hardest (with highest loss value) samples.
\item To train the network in mini-batches as usual on hardest samples.
\item To increase the difficulty of the augmentation and go to beginning.
\end{enumerate}

\par The last component to train the network successfully is a strong data augmentation with the progressively increased difficulty. The best choice is to use random erasing augmentation \cite{DBLP:journals/corr/abs-1708-04896}  in addition to standard horizontal flip and random crop methods.


\begin{figure*}[th]
\centering
\includegraphics[width=0.9\textwidth]{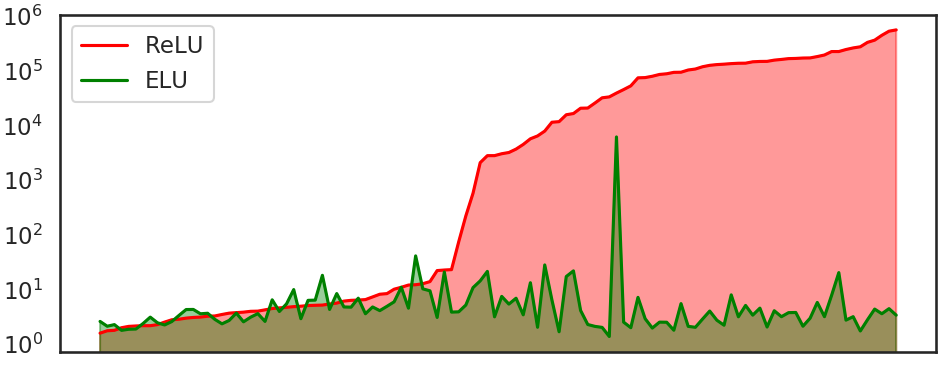}
\caption{Comparison of ratio $w_{max} / w_{min}$ of learnt model weights for the different activation functions. Filters are ordered according the the ReLU ratios.}
\label{fig:relu-vs-elu}
\end{figure*}

\section{Experimented Result}

\subsection{Data}
\lettrine[nindent=0em,lines=3]{T}{}o evaluate the proposed solution we use the Market-1501 dataset \cite{zheng2015scalable}. It is a benchmark for person re-identification purposes with images from 6 cameras of different resolutions. It was annotated with the 1501 identities: 751 among which are used for training and 750 are used for testing. The training set contains 12936 images with 3368 query images. The gallery set is composed of images from the 750 test identities and of distractor images, 19732 images in total. The most common and useful evaluation scenario is a single query image.

\subsection{Metrics}
\lettrine[nindent=0em,lines=3]{W}{}e follow standard procedure and report the mean  average precision over all queries (mAP) and the cumulative matching curve (CMC) at rank-1 using the evaluation codes provided by the benchmark.

\par It's worth saying that there are some techniques to improve the final result in both metrics. The first common method is to estimate the embedding for the original and flipped images and then concatenate them into a single one (including additional normalization step to use with cosine similarity measure). In our opinion it is not an honest way to improve the accuracy because it doubles the computation time. Unfortunately some authors don't report the result with a mark that flipping is used. However to be able to go with that approach we report results including horizontal flipping metric.

\par The second method is based on using re-ranking (RK) techniques \cite{DBLP:journals/corr/ZhongZCL17}. In other words it is direct optimization over comparable metrics. We report result with RK too.

\subsection{Ablation study}

\lettrine[nindent=0em,lines=3]{A}{}s it was announced earlier we first compare the backbone implementations with different activation functions. Our main message in this paper is that widely used ReLU activation is not a proper one that leads to uprising of some problems. To prove it we measure the ratio $w_{max} / w_{min}$ between absolute values of filter weights for each convolution layer in network. On Figure \ref{fig:relu-vs-elu} you can find this ratios for both ReLU and ELU activation functions. High value of ratio means that there are invalid filters on the current level. As it can be seen the network trained with ReLU have more than half of noisy filters which usually is pruned for model compression purposes. Another picture gives us the result of using the ELU activation function -- significant part of filters is still useful and no capacity reduction is observed. Due to low final quality of model with ReLU activation all the next experiments are performed with ELU.

\par Table \ref{table:abstudy} shows ablation of study experiments. The initial point of our experiments is training on our dataset with AM-Softmax loss only. Generally speaking this approach should beat SOTA results with general-purpose backbone. But in our case we are very limited in the model capacity and default training is failed. In other words the task to train the lightweight but accurate person re-identification network is really challenging.

\begin{table}[hb]
\begin{tabular}{l|cc}
\multicolumn{1}{c|}{\multirow{2}{*}{\textbf{Method}}} & \multicolumn{2}{c}{\textbf{Market-1501}} \\
\multicolumn{1}{c|}{} & \textbf{rank@1} & \textbf{mAP} \\ \hline
AM-Softmax & 78.00 & 60.74 \\
+ HSM & 79.07 & 57.88 \\
+ Center loss & 81.53 & 60.42 \\
+ Disabled dropout & 85.24 & 65.94 \\
+ Push loss & 87.11 & 70.95 \\
+ GlobPush loss & 88.69 & 73.40 \\
+ Smart margins & 90.20 & 78.80 \\
+ Weighted HSM & 91.66 & 81.63 \\
+ Increased resolution & 92.37 & 82.53
\end{tabular}
\caption{Ablation study on Marlet-1501 dataset. HSM -- hard sample mining procedure.}
\label{table:abstudy}
\end{table}

\begin{table*}[b]
\begin{tabular}{l|cc|ccc}
\multicolumn{1}{c|}{\multirow{2}{*}{\textbf{Method}}} & \multicolumn{2}{c}{\textbf{Market-1501}} & \multirow{2}{*}{\textbf{GFLOPs}} & \multirow{2}{*}{\textbf{MParams}} & \multirow{2}{*}{\textbf{FPS}} \\
\multicolumn{1}{c|}{} & \textbf{rank@1} & \textbf{mAP} &  &  &  \\ \hline
GP-ReID \cite{DBLP:journals/corr/abs-1801-05339} & 92.2 & 81.2 & 8 & 24.66 & 64 \\
Deep-Person \cite{DBLP:journals/corr/abs-1711-10658} & 92.3 & 79.5 & 8 & 24.66 & 64 \\
PCB \cite{DBLP:journals/corr/abs-1711-09349} & 92.4 & 77.3 & 8 & 24.66 & 64 \\
PCB+RPP \cite{DBLP:journals/corr/abs-1711-09349} & 93.8 & 81.6 & 8 & 24.66 & 64 \\
HPM (flip) \cite{DBLP:journals/corr/abs-1804-05275} & 94.2 & 82.7 & $2 \times 8$ & 24.66 & 32 \\
MGN (flip) \cite{DBLP:journals/corr/abs-1804-01438} & 95.7 & 86.9 & $2 \times 24$ & 68.75 & 16 \\ \hline
Our (light) & 91.7 & 81.6 & 0.12 & 0.81 & 923 \\
Our (strong) & 92.4 & 82.5 & 0.58 & 0.81 & 268 \\
Our (strong, flip) & 92.5 & 83.1 & $2 \times 0.58$ & 0.81 & 134 \\ \hline
MGN (RK) \cite{DBLP:journals/corr/abs-1804-01438} & 96.6 & 94.2 & $2 \times 24$ & 68.75 & 16 \\
Our (strong, RK) & 93.1 & 91.1 & 0.58 & 0.81 & 268
\end{tabular}
\caption{Comparison with state of the art solutions on Market-1501 dataset. Performance results (Frames Per Second) are measured with OpenVINO on Intel Core i7-6700K CPU@2.90GHz}
\label{table:results}
\end{table*}

\par The first step to improve the baseline is to tackle with the data imbalance problem. As it was mentioned earlier in this paper we use the hard sample mining (HSM) procedure (see the description of the used method above). In the first experiment the AM-Softmax loss value is used to order the samples only (instead of step $b$). The impact on metrics is not significant but it allows us not to think more about possible over-fitting due to training on plain samples.

\par During next steps we dive into the manifold learning approach by introducing different local structure losses: Center, Push and GlobPush. Each step gives us the following improvement in both metrics. The most significant impact is achieved after using the smart margins for the Push and GlobPush losses. Moreover as it can be expected smart margins mostly affect the mAP metric which reflects the orderliness of the learnt manifold.

\par It is worth noting that our concerns about the limited model capacity due to using dropout regularization are confirmed and the strategy to disable this type of regularization on the late iterations brings us significant leap in accuracy for both metrics.

\par The very last attempt to increase the metrics is to make the sample mining procedure more flexible (more \textit{smart} to consider the sample complexity for different-level losses) by mixing multiple losses into the ranking criterion. It allows us to improve the mAP metric mostly.

\par To be able to align with other state of the art solutions we should also test different input resolutions. As it was shown in paper \cite{DBLP:journals/corr/abs-1801-05339} the plain increasing of input size can bring significant leap in accuracy. For our model the main input resolution is $160 \times 64$. We also tested the higher $384 \times 128$ resolution and as it can be seen the result is slightly better but with expected slowdown in the inference time (both model variants can be found in Intel$^{\textregistered}$ OpenVINO\texttrademark toolkit).

\subsection{Comparison with the state of the art}

\lettrine[nindent=0em,lines=3]{T}{}able \ref{table:results} compares the proposed solution to the state of the art approaches. Our approach without using multibranching \cite{DBLP:journals/corr/abs-1804-01438} or merging embeddings from the different levels \cite{DBLP:journals/corr/abs-1804-05275} achieves well enough accuracy but significantly outperforms in the inference time more than one order of magnitude. It happens due to our lightweight backbone RMNet instead of the widely used ResNet-50 architecture.

\par The proposed combination of loss functions and the training strategy allows us to achieve the comparable results even when our model has significantly less number of parameters (0.81 vs 25 MParams). Moreover our solution is in top-3 by rank@1 metric and in top-2 by the mAP metric.

\par To measure the model performance we use publicly available OpenVINO toolkit and run experiments on Intel Core i7-6700K CPU. We significantly outperforms other solutions by the Frame per Second (FPS) metric. It is worth saying that person re-identification method can be referred to real-time solutions if it's able to perform several pairwise comparisons on each frame from the input stream in the real-time mode. For example using our faster solution (\textit{light}, 923 fps) we can process about 30 persons on each frame in real-time. No one other state of the art is able to do that on the same quality.

\section{Conclusion}

\lettrine[nindent=0em,lines=3]{I}{}n this paper we have proposed the novel lightweight backbone (RMNet) and set of training practices to tackle with the person re-identification problem. We have demonstrated that our solution is close to state of the art approaches but significantly outperforms them by the inference time. We presume that our work gives new breath to the lightweight solution development for the wide range of applications by direct designing of task-specific networks.


\bibliographystyle{splncs}
\bibliography{egbib}


\end{document}